\pdfoutput=1

\documentclass[11pt]{article}
\usepackage{emnlp2021}
\usepackage{emnlp2021}
\usepackage{times}
\usepackage{latexsym}
\usepackage{url}
\usepackage{examples}
\usepackage{multirow}
\usepackage{graphicx}
\usepackage{amsmath}
\usepackage[para,online,flushleft]{threeparttable}
\usepackage{tablefootnote}
\usepackage[toc,page]{appendix}

\usepackage{microtype}

\title{
End-to-end Neural Information Status Classification
}

\author{
Yufang Hou \\
  IBM Research Europe\\ Dublin, Ireland\\ {\tt yhou@ie.ibm.com}\\
   }

\date{}
\begin{document}

\maketitle

\begin{abstract}
Most previous studies on information status (IS) classification and bridging anaphora recognition assume that the gold mention or syntactic tree information is given \cite{houyufang13b,houyufang16,roesiger18a,hou-2020-fine,yu-poesio-2020-multitask}. 
In this paper, we propose an end-to-end neural approach for information status classification. Our approach consists of a 
 mention extraction component and an information status assignment component. During the inference time, our system takes a raw text as the input and generates mentions together with their information status. On the ISNotes corpus \cite{markert12}, we show that our information status assignment component achieves new state-of-the-art results on fine-grained IS classification based on gold mentions. Furthermore, our system performs significantly better than other baselines  for both mention extraction and fine-grained IS classification in the end-to-end setting. Finally, we apply our system on BASHI \cite{roesiger18b} and SciCorp \cite{roesiger-2016-scicorp} to recognize referential bridging anaphora. We find that our end-to-end system trained on ISNotes achieves competitive results on bridging anaphora recognition compared to the previous state-of-the-art system that 
relies on syntactic information and is trained on the in-domain datasets \cite{yu-poesio-2020-multitask}.

\end{abstract}

\section{Introduction}
\label{sec:intro}


\emph{Information status} (IS henceforth) studies how discourse entities are referred to in a text \cite{halliday67,prince81a,nissim04,markert12} and is related to a wide range of discourse phenomena, such as \emph{coreference} \cite{ng-2010-supervised}, \emph{bridging} \cite{clarkherberth75}, and \emph{comparative anaphora} \cite{modjeska-etal-2003-using}. In general, IS reflects the accessibility of a discourse entity based on the evolving discourse context and the speaker’s assumption about the hearer’s knowledge and beliefs.  

\begin{figure*}[t]
\begin{center}
\includegraphics[width=0.97\textwidth]{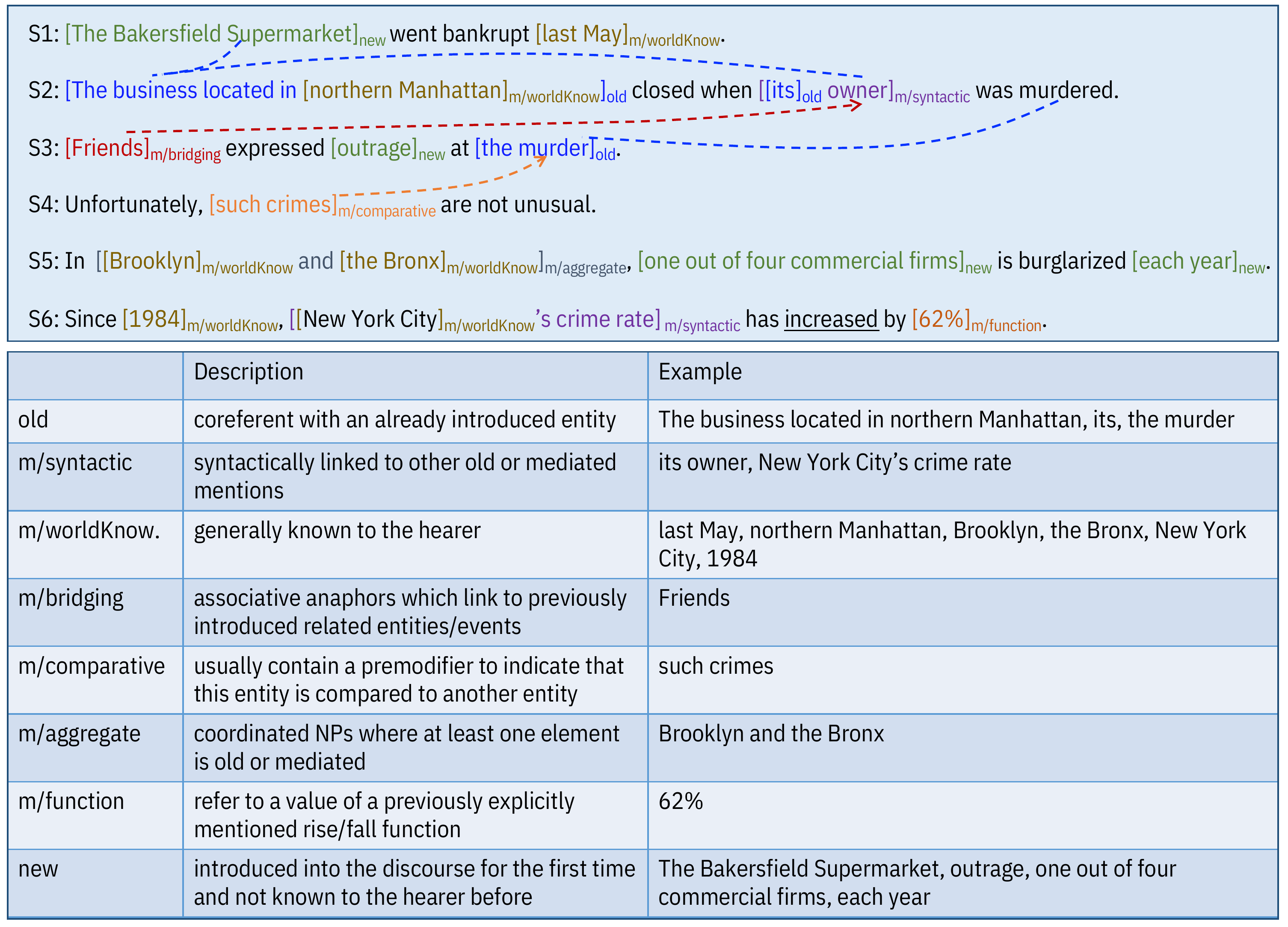}
\end{center}
\caption{Information status categories and examples. IS definitions are from \newcite{hou-2020-fine}. Examples are adapted from a news article in ISNotes.}
\label{fig:definition}
\end{figure*}

Based on \newcite{prince92} and \newcite{nissim04}, \newcite{markert12} proposed an IS scheme for written English which consists of eight IS categories. According to \newcite{markert12}, a \emph{mention} is a discourse entity that carries information status. An entity is \emph{old} if it refers to the same entity that has been mentioned previously.  \emph{Mediated} entities are discourse-new and hearer-old. They have not been introduced in the discourse directly, but are inferrable from previously mentioned entities, or generally known to the hearer. Finally, an entity is \emph{new} if it is introduced into the discourse for the first time and the hearer cannot infer it from previously mentioned entities. Figure \ref{fig:definition} explains each IS category with examples from a short text.

Unlike coreference resolution, most previous work on IS classification and bridging resolution assumes that the gold mention information 
is given \cite{rahman12,houyufang13b,houyufang16,roesiger18a,hou-2020-fine,yu-poesio-2020-multitask}. On a few corpora where gold mentions are not annotated (e.g., BASHI), researchers rely on syntactic parsers to extract 
mentions \cite{houyufang18b,yu-poesio-2020-multitask}. In this paper, we demonstrate that mentions can be reliably extracted without relying on syntactic information. Note that
 we extract all \emph{singleton} as well as \emph{non-singleton} mentions and assign information status to them. 
This is different from most previous work on coreference resolution \cite{leeheeyoung11,clark-manning-2016-improving,lee-etal-2017-end} where the mention extraction component is focused on identifying non-singleton mentions only.

Our system consists of two models. The first model (\emph{MenExt}) 
uses context-dependent boundary representations to 
extract mentions of discourse entities including \emph{singleton} mentions. 
Unlike \newcite{lee-etal-2017-end}, our model does not have the constraint of the maximum mention length, and it reasons over the space of all spans during the inference time. The second model (\emph{ISAssign}) assigns information status to the extracted mentions. 
We use a mention's boundary representations in its corresponding context to predict its information status. The context information is adapted from \newcite{hou-2020-fine} which is syntactic-agnostic.

For information status classification based on gold mentions, our second model \emph{ISAssign} significantly outperforms the existing state-of-the-art model \cite{hou-2020-fine} on the ISNotes benchmark based on the same context information. In the end-to-end setting, our  system achieves strong results for both mention extraction and IS classification compared to other baselines on ISNotes. We further demonstrate the usefulness of our system  by applying it to identify referential bridging anaphora on BASHI \cite{roesiger18b} and SciCorp \cite{roesiger-2016-scicorp}. We find that our end-to-end system trained on ISNotes achieves competitive 
 results for bridging anaphora recognition compared to a recent approach \cite{yu-poesio-2020-multitask} which uses syntactic tree information for mention extraction and is trained on the in-domain datasets. We will release code and the trained models at \url{https://github.com/IBM/bridging-resolution/}.
 

\section{Related Work}
\paragraph{Information Status Classification.} Previous work on information status classification all assumes that the gold mention information is available \cite{markert12,rahman12,cahill12,houyufang13b,houyufang16,hou-2020-fine}.
In this work, we remove this constraint and propose a system to tackle the task in an end-to-end manner. 

\begin{figure*}[t]
\begin{center}
\includegraphics[width=1.0\textwidth]{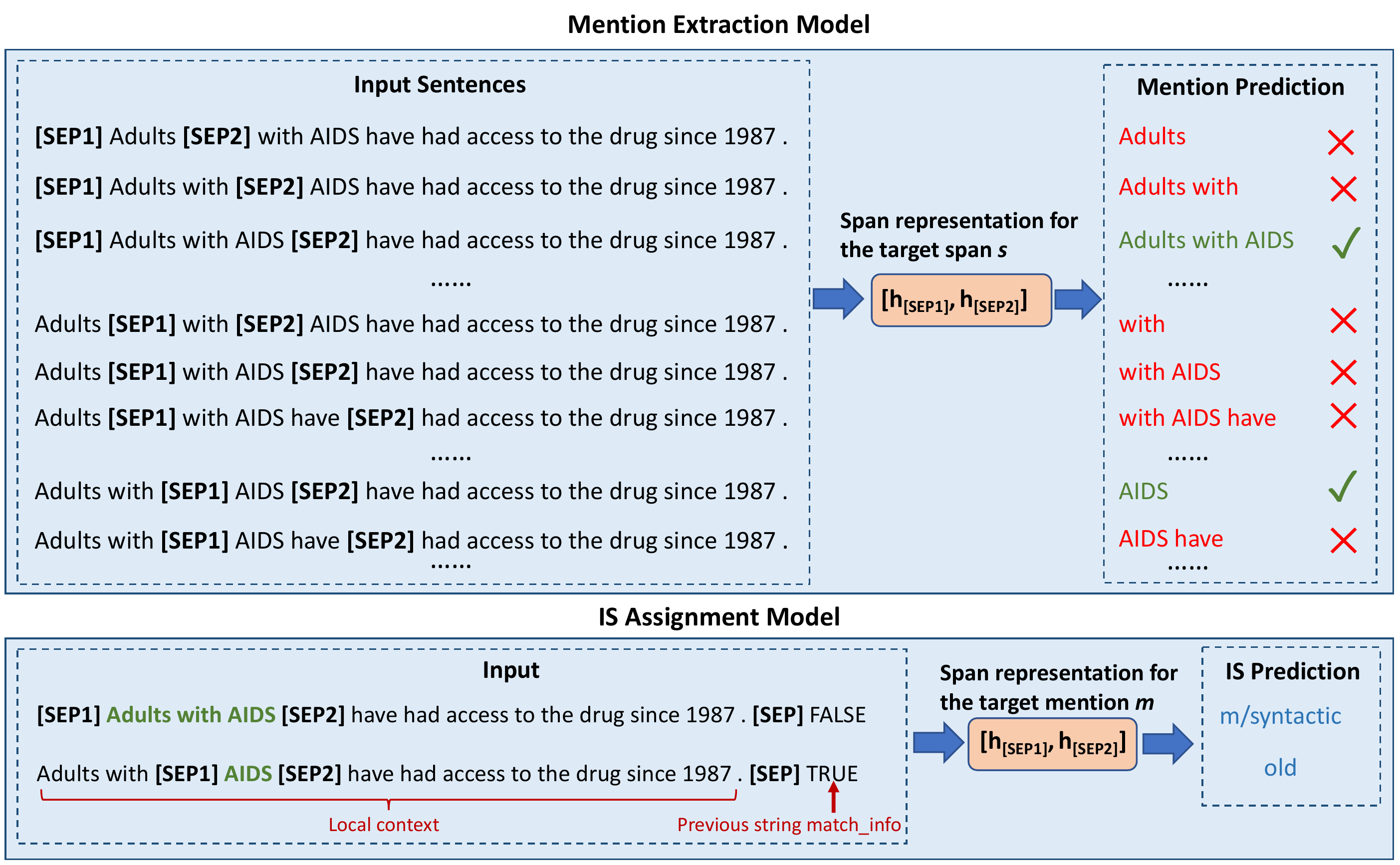}
\end{center}
\caption{System architecture of our proposed approach for end-to-end IS classification.}
\label{fig:model}
\end{figure*}

\paragraph{Bridging Anaphora Recognition.}
Most previous work on bridging focuses on selecting the correct antecedents for gold bridging anaphors \cite{poesio04d,houyufang13a,houyufang18b,houyufang20a}. 
As a sub-task of bridging resolution \cite{houyufang18c}, bridging anaphora recognition requires a system to identify all bridging anaphors in a text. Some previous work models this task as part of IS classification problem \cite{houyufang13b,hou-2020-fine}.
  \newcite{yu-poesio-2020-multitask} propose a 
multi-task framework for bridging resolution and report results for bridging anaphora recognition on a wide range of bridging corpora that follow different bridging definitions. 
All these studies are mainly based on gold mentions. On a few corpora (BASHI and SciCorp) in which mentions are not annotated, gold or predicted syntactic information is explored to extract ``NPs'' as the mentions.
In this work, we focus on identifying bridging anaphors on the three corpora (i.e., ISNotes, BASHI, and SciCorp) that contain anaphoric referential bridging annotations.\footnote{\newcite{roesiger18a} call this type of bridging as ``referential bridging'' in which bridging anaphors are truly anaphoric. This is different from bridging annotations in ARRAU \cite{poesio18} where most bridging links are ``non-anaphoric referential bridging pairs", such as \emph{Europe--Spain}.} Our system does not rely on any gold mention or syntactic information and works well across different domains.



\paragraph{Mention Extraction.} 
Mention extraction is an important task for a lot of down-streaming tasks, such as \emph{relation extraction} \cite{mintz-etal-2009-distant}, \emph{entity linking} \cite{kolitsas-etal-2018-end}, and \emph{coreference resolution} \cite{leeheeyoung11}. 
Note that the meaning of ``mention'' varies across different tasks.
 Our definition for mentions is close to the one used in coreference resolution.
However, mention detection in coreference resolution usually focuses on non-singletons only. In our work, mentions are referential discourse entities which include both singleton and non-singleton entities. Furthermore, IS classification is also related to ``anaphorocity detection'' in coreference resolution \cite{wiseman-etal-2015-learning} as some \emph{old} mentions are anaphors in coreference. 

 
%
%


\section{Method}
In this section, we describe our system for end-to-end information status classification in detail. Figure \ref{fig:model} illustrates how the system extracts mentions from the input sentences (\emph{Mention Extraction Model}) and assigns information status for the predicted mentions (\emph{IS Assignment Model}). 


\subsection{Mention Extraction Model}
\label{sec:mentionExt}
We formulate the task of mention extraction as a binary classification problem for every possible span in a sentence $sent$. 
Specifically, for a span $s$ containing one or multiple words, we insert two special tokens ``[SEP1]'' and ``[SEP2]'' immediately before and after $s$, respectively. This gives us a new sentence ${sent}'$. We then use a transformer encoder to encode the new sentence ${sent}'$ and 
concatenate the hidden states of ``[SEP1]'' and ``[SEP2]'' in the last layer to generate the contextual representation for the span $s$ (i.e., [$h_{[SEP1]}$, $h_{[SEP2]}$]). The probability of $s$ being a valid mention is calculated by a softmax function over the output layer based on the span representation ``[$h_{[SEP1]}$, $h_{[SEP2]}$]''. 

\paragraph{Training.}
Training data for mention extraction is generated by considering spans up to $L$ words for all sentences. For a sentence containing $N$ words, we generate the corresponding training instances by collecting spans that start from each word up to $L$ words (illustrated in ``Input Sentences'' in Figure \ref{fig:model}). 
The size of the training instances generated for each sentence will be $\mathcal{O}(NL)$ if $L<N$. To make training more efficient, following \newcite{lee-etal-2017-end}, we set $L= 10$.\footnote{ \newcite{lee-etal-2017-end} 
reported that mentions that exceed the maximum mention
width of ten words only account for less than 2\% of the
training mentions in OntoNotes. Note that these mentions are all non-singletons.
In ISNotes, we found that 10\% of the mentions contain more than ten words and most of them (80\%) are singletons.} 

\paragraph{Inference.}
During the inference stage, for a given sentence $s$ that contains $N$ words, we apply the learned model to all spans ($\mathcal{O}(N^2)$) without any pruning. The spans that belong to the positive class are extracted as the predicted mentions from this sentence.

It is worth noting that previous work on coreference resolution usually uses aggressive pruning strategies to extract non-singleton mentions \cite{lee-etal-2017-end,wu-etal-2020-corefqa}.
 In our work, the mention extraction model is trained using all spans up to ten words which include both singleton and non-singleton mentions as the positive instances. Although this filters out a small portion of long mentions, we think the fact that our training data contains the signals of mention/non-mentions for all spans up to ten words is sufficient for the model to learn the differences between mentions and non-mentions. Therefore, we do not apply any pruning during inference. 


\subsection{IS Assignment Model}
\label{sec:isAssignment}
We model IS assignment as a multi-class classification problem for every predicted mention in a sentence $\emph{sent}$. More specifically, for a predicted mention $m$ in $\emph{sent}$, we insert two special tokens ``[SEP1]'' and ``[SEP2]'' immediately before and after $m$, respectively. We treat this new sentence as the local context to predict the IS category for $m$. Furthermore, we add an additional token which indicates whether the current mention $m$ has the same string with any previous mentions in the text. \newcite{hou-2020-fine} found that this additional previous context information is important for predicting \emph{old} mentions. As illustrated in Figure \ref{fig:model} (IS Assignment Model), we generate the input for the model by concatenating the local context and the additional token using the special token ``[SEP]''. Similar to the mention extraction model architecture, we use another transformer encoder to encode the input and generate the contextual representation ([$h_{[SEP1]}$, $h_{[SEP2]}$]) for the mention $m$ by concatenating the hidden states of ``[SEP1]'' and ``[SEP2]'' in the last layer. Finally, 
we use a softmax function to predict the IS category for $m$ based on its contextual representation ``[$h_{[SEP1]}$, $h_{[SEP2]}$]''. 

Note that our model is different from the one proposed in \newcite{hou-2020-fine}. \newcite{hou-2020-fine} treats IS assignment as a sentence classification task by designing a special sequence for each mention. The author applies BERT to encode the sequence and uses the hidden state of the ``[CLS]'' token for prediction. Although this model achieves a great improvement compared to previous approaches on IS classification on ISNotes, we think that our mention representation based on ``[SEP1]'' and ``[SEP2]'' is more accurate compared to the ``[CLS]'' token because the latter does not capture a mention's position information in its local context.\footnote{For both mention extraction and IS assignment, we have tried to extract representation for a mention based on its first token and last token. In our experiments, we found that using the special tokens (\emph{[SEP1]}/\emph{[SEP2]}) leads to better results.} 


\paragraph{Training and Inference.}
The training dataset is generated based on the gold mentions. During inference, we applied the trained model to assign information status to the predicted mentions or the gold mentions in the testing dataset.

\subsection{Model Parameters}
Our system is developed based on Huggingface Transformers \cite{wolf-etal-2020-transformers}. For mention extraction, we train the model for one epoch with a learning rate of $1e-5$\footnote{In practice, we found that a higher learning rate leads to unstable models for mention extraction. This corresponds to the observations from \newcite{mosbach2020stability} that small learning rates can avoid vanishing gradients early in training when fine-tuning transformer-based masked language models.} and a batch size of 32. For IS assignment, the model is trained for three epochs with a learning rate of $3e-5$ and a batch size of 32. Both models are initialized using pre-trained RoBERTa$_{LARGE}$ contextual embeddings \cite{liu2019roberta}. During training and testing, the maximum text length is set to 128 tokens.\footnote{The maximum text length is set to 256 tokens when applying the system to SciCorp to recognize bridging anaphors. This is because SciCorp contains a few long sentences due to the noise from sentence splitting.}


\section{Experiments}
\subsection{Datasets}
We use three datasets for experiments. The first dataset ISNotes \cite{markert12} contains 50 texts with 10,980 mentions from the World Street Journal (WSJ) portion of the OntoNotes corpus. Each mention is assigned to one of the eight IS categories described in Figure \ref{fig:definition}. Table \ref{tab:is} shows the IS distribution in ISNotes.

For mention extraction and IS assignment, we mainly test our approach on ISNotes since it contains mention annotations for all mentions including both singletons and non-singletons.  

Additionally, we apply our model trained on ISNotes to identify bridging anaphors in BASHI \cite{roesiger18b} and SciCorp \cite{roesiger-2016-scicorp} in which bridging anaphors are truly anaphoric\footnote{\newcite{roesiger18a} call this type of bridging as ``\emph{referential bridging}''.}.  In general, bridging anaphora recognition is a relatively less explored research area, especially in the end-to-end setting. 

BASHI contains 459  bridging anaphors from 50 WSJ texts which are different from the ones in ISNotes. Note that BASHI does not contain mention annotations and its bridging anaphors also include comparative anaphors which are treated as a separate IS category in ISNotes.

SciCorp  contains 1,366 bridging anaphors from 14 full English scientific papers from two disciplines (i.e., genetics and computational linguistics). Note that bridging anaphors in this corpus are restricted to definite noun phrases only. Following \newcite{roesiger18a}, we filter out bridging anaphors that are ``\emph{containing inferrable}'' \cite{prince81a}, such as ``\emph{their interest}'' or ``\emph{our aim}''. These mentions are ``\emph{mediated/syntactic}'' in ISNotes.


\begin{table}[t]
\begin{center}
\begin{tabular}{lrr}
Mentions & &10,980 \\ \hline\hline
old & & 3,237 \\ \hline
mediated & & 3,708\\ \hline
& syntactic & 1,592\\
& world knowledge & 924\\
& bridging & 663\\
& comparative & 253\\ 
& aggregate & 211\\
& func & 65\\
 \hline
new & & 4,035\\
\hline
\end{tabular}
\end{center}
\caption{\label{tab:is}  IS distribution in ISNotes.}
\end{table}

\subsection{Experimental Setup}
On ISNotes, all experiments are performed using 10-fold cross-validation on documents. 
For mention extraction, we report \emph{Recall}, \emph{Precision} and \emph{F-score} based on exact match on gold mentions. 
For IS assignment, we report \emph{Recall}, \emph{Precision} and \emph{F-score} per IS category. In the end-to-end setting, a prediction is counted as correct if it matches with the boundaries of a gold mention and has the same IS type as the gold mention. 

\begin{table}[t]
\begin{center}
\begin{tabular}{|l|c|c|c|}
\hline
 &R&P&F\\  \hline
  \multicolumn{4}{|c|}{Baselines}\\ \hline
\emph{syntactic NPs}& 78.0&67.1&72.1\\ \hline
\emph{Lee et al. (2011)}&74.2&70.8&72.5\\ \hline
\emph{Yu et al. (2020)}&88.7&86.6&87.7\\ \hline
  \multicolumn{4}{|c|}{Our model}\\ \hline
\emph{our model \begin{small}(test $L=10$)\end{small}}&83.5&\textbf{92.4}&87.7\\ \hline
\emph{our model}&\textbf{92.8}&91.5&\textbf{92.1}\\ \hline
\end{tabular}
\end{center}
\caption{\label{tab:mentionExt} Results of our mention extraction model compared to the baselines on ISNotes. The differences between our model and the baselines are statistically significant at p$<$0.01 using randomization test.}
\end{table}

\subsection{Results for Mention Extraction on ISNotes}
\label{sec:mentionExtResult}
For mention extraction, we compare our method with three baselines. 
The first baseline (\emph{syntactic NPs}) is simply to extract all NPs from the automatically parsed syntactic trees as mentions.  \newcite{yu-poesio-2020-multitask} use this strategy to predict mentions 
for BASHI and SciCorp because gold mentions are not annotated in these two datasets.
We use the Stanford CoreNLP toolkit \cite{manning-etal-2014-stanford} to obtain the  syntactic trees.
The second baseline (\emph{Lee et al. (2011)}) is the mention detection component from \newcite{leeheeyoung11} which is widely used in various coreference resolution systems. It is a rule-based system based on automatically parsed trees.
The third baseline (\emph{Yu et al. (2020)}) is the best neural mention detection approach proposed in \newcite{yu-etal-2020-neural} which uses biaffine attention over a bi-directional LSTM to predict mentions.  

\begin{table*}[t]
\begin{small}
\begin{center}
\begin{tabular}{|p{2.6cm}|p{0.8cm}p{0.8cm}p{0.8cm}|p{0.8cm}p{0.8cm}p{0.8cm}|p{0.8cm}p{0.8cm}p{0.8cm}|}
\hline
  &  \multicolumn{3}{c|}{\emph{baseline}} &\multicolumn{6}{c|}{\emph{this work}} \\ 
&  \multicolumn{3}{c|}{\emph{self-attention with}} &\multicolumn{3}{c|}{\emph{baseline with}} & \multicolumn{3}{c|}{\emph{our model}} \\ 
 &  \multicolumn{3}{c|}{\emph{BERT$_{LARGE}$}} &\multicolumn{3}{c|}{\emph{RoBERTa$_{LARGE}$}} & \multicolumn{3}{c|}{\emph{RoBERTa$_{LARGE}$}} \\ 
 & R & P & F & R & P & F& R & P & F\\ \hline
\hline
 &&&&&&&&& \\
 {old} & 88.4&90.0&89.2&89.0&92.0&\textbf{90.5}&88.8&91.8&90.3\\
 {m/worldKnow.} & 77.7&79.5&78.6&78.0&80.9&\textbf{79.4}&79.2&79.4&79.3\\
 {m/syntactic}&83.7&81.1&82.4&85.7&81.8&83.7&84.7&83.5&\textbf{84.1}\\
 {m/aggregate} & 80.1&79.3&\textbf{79.7}&77.7&75.9&76.8&76.8&77.5&77.1\\
 {m/function}& 73.4&85.5&79.0&71.9&80.7&76.0&90.6&81.7&\textbf{85.9}\\
 {m/comparative} &90.5&86.7&\textbf{88.6}&78.7&85.4&81.9&87.7&88.1&87.9\\
 {m/bridging} & 51.0&54.5&52.7&47.8&53.5&50.5&54.1&59.9&\textbf{56.9}\\
{new}&86.6&85.2&85.9&88.2&84.9&86.5&88.8&85.8&\textbf{87.3}\\
 \hline
acc &\multicolumn{3}{c|}{83.7} & \multicolumn{3}{c|}{84.3}
 & \multicolumn{3}{c|}{\textbf{85.1}}\\
\hline
\end{tabular}
\end{center}
\end{small}
\caption{Results of the IS assignment model compared to the previous state-of-the-art approach on ISNotes. Bolded scores indicate the best performance for each IS class. The difference between \emph{our model with RoBERTa$_{LARGE}$} and the baseline is statistically significant at p$<$0.01 using randomization test.}
\label{tab:isnotes-goldmention}
\end{table*}

Table \ref{tab:mentionExt} shows the results of our model on mention extraction compared to the three baselines. 
Our model outperforms the three baselines in all metrics. In particular, our model achieves an F-score of 92.1, 
obtaining a 4.4 absolute improvement in F-score compared to the baseline \emph{Yu et al. (2020)}.\footnote{\newcite{yu-etal-2020-neural} report 87.9 (recall), 89.7  (precision) and 88.8 (F-score) for mention detection on the CRAC testing dataset in the “High F1” setting. Our mention extraction model trained on the CRAC training dataset achieves better results on the same test set: 91.1 (recall), 90.4 (precision) and 90.8 (F-score).}

Table \ref{tab:mentionExt}  also reports the results of our model using $L=10$ as the pruning strategy during inference. That is, we filter out predicted mentions which contain more than ten words. This results in a 4.4 F-score drop in performance,  
which verifies our assumption in Section \ref{sec:mentionExt} that our model can learn the differences between mentions and non-mentions well from the signals of all spans up to ten words and we do not need to apply pruning during inference.

\paragraph{Inference Efficiency.} During the inference stage for mention extraction, previous studies on coreference resolution normally set the maximum mention length to 10 tokens \cite{lee-etal-2017-end}. \newcite{yu-etal-2020-neural} set this number to 30. We implemented both options as our baselines (see Table \ref{tab:mentionExt}). For a sentence containing $k$ words, our inference algorithm will have to classify $k(k+1)/2$ spans, which leads to a time complexity of $O(k^2)$. The other two options will have a time complexity of $O(k*10)$ and $O(k*30)$, respectively. The average sentence length on ISNotes is 24.3 tokens. Among all sentences, 10.5\% have less than 10 tokens, and 73\% have less than 30 tokens. Normally $k$ would not be a large number. On a sentence containing 100 tokens, our algorithm is 5 times slower than the pruning with $L=10$, and 1.7 times slower than the pruning with $L=30$. In practice, we have tested our algorithm on scientific papers using a simple heuristic to filter out non-mentions during inference. More specifically, we do not perform inference for spans starting with verbs, punctuation or prepositions using a dictionary. This greatly helps us to speed up the inference process. In addition, one main goal of our work is to test the performance of this algorithm without adding any pruning constraints. As a result, 
it is interesting to see that the model can effectively filter out long non-mentions by itself.

\subsection{Results for IS Classification on ISNotes Based on Gold Mentions}
In this section, we compare our IS assignment model to the previous state-of-the-art approach \cite{hou-2020-fine} using gold mentions as input.  
Note that the baseline 
model proposed by \newcite{hou-2020-fine} treats IS classification as a sentence classification task based on carefully designed ``pseudo sentences''. It is fine-tuned on \emph{BERT$_{LARGE}$} and uses the hidden state of the ``[CLS]'' token for prediction.\footnote{Please check Appendix \ref{sec:moreisresults} for more results of IS classification on ISNotes from previous approaches.} 

We find that changing \emph{BERT$_{LARGE}$} to \emph{RoBERTa$_{LARGE}$}  improves the overall accuracy by 0.6\%  (see \emph{baseline with \emph{RoBERTa$_{LARGE}$}} in Table \ref{tab:isnotes-goldmention}). 
Our model further improves the accuracy by 0.8\% (85.1 vs. 84.3 in Table \ref{tab:isnotes-goldmention}), which demonstrates that the mention representation based on the boundary embeddings in our model is more accurate than the baseline model. In addition, it seems that our mention representation is especially effective for detecting bridging anaphors and \emph{mediated/function} mentions. 


\begin{table*}[t]
\begin{small}
\begin{center}
\begin{tabular}{|p{2.15cm}|p{0.69cm}p{0.69cm}p{0.69cm}|p{0.69cm}p{0.69cm}p{0.69cm}|p{0.69cm}p{0.69cm}p{0.69cm}|p{0.69cm}p{0.69cm}p{0.69cm}|}
\hline
  &  \multicolumn{9}{c|}{\emph{baselines}} &\multicolumn{3}{c|}{\emph{this work}} \\ 
&  \multicolumn{3}{c|}{\emph{NPs from predicted}} &\multicolumn{3}{c|}{\emph{system mentions}}& \multicolumn{3}{c|}{\emph{system mentions}}& \multicolumn{3}{c|}{\emph{our mention}} \\ 
 &  \multicolumn{3}{c|}{syntactic trees} &\multicolumn{3}{c|}{\emph{Lee et al. (2011)}} &\multicolumn{3}{c|}{\emph{Yu et al. (2020)}} & \multicolumn{3}{c|}{\emph{extraction model}} \\ 
 & R & P & F & R & P & F&R & P & F& R & P & F\\ \hline
\hline
 &&&&&&&&&&&& \\
 {old} & 66.4&77.3&71.4&79.8&81.4&80.6&83.2&86.6&84.9&85.0&89.1&\textbf{87.0}\\
 {m/worldKnow.} & 49.8&58.9&54.0&65.0&61.3&63.1&72.4&69.3&70.8&74.9&70.8&\textbf{72.8}\\
 {m/syntactic}&67.0&53.9&59.8&57.0&60.8&58.9&77.4&70.2&73.6&80.4&76.3&\textbf{78.3}\\
 {m/aggregate} & 60.7&61.5&61.1&28.0&31.9&29.8&61.1&63.2&62.2&69.7&72.1&\textbf{70.8}\\
 {m/function}& 76.6&63.6&69.5&70.3&50.6&58.8&73.4&73.4&73.4&81.2&81.2&\textbf{81.2}\\
 {m/comparative} &73.5&56.9&64.1&52.2&55.2&53.7&74.3&72.3&73.3&82.2&79.7&\textbf{80.9}\\
 {m/bridging} & 50.2&42.1&45.8&42.2&44.5&43.3&51.6&53.1&52.3&51.3&53.5&\textbf{52.4}\\
{new}&70.8&49.1&58.0&58.2&49.8&53.7&74.1&70.6&72.3&79.0&75.4&\textbf{77.1}\\
 \hline
overall &65.8&56.7&60.9&63.4&60.5&61.9&75.5&73.8&74.6&78.9&77.8&\textbf{78.3}\\
\hline
\end{tabular}
\end{center}
\end{small}
\caption{Results of IS classification on ISNotes in the end-to-end setting using different mention extraction approaches. Bolded scores indicate the best performance for each IS class. The differences between our system and other baselines are statistically significant at p$<$0.01 using randomization test.}
\label{tab:isnotes-extractedmention}
\end{table*}

\subsection{Results for IS Classification on ISNotes Based on Extracted Mentions}
Previous studies on IS classification only focus on gold mentions \cite{markert12,rahman12,cahill12,houyufang13b,hou-2020-fine}.
It is unclear how these approaches will perform in realistic scenarios where gold mentions are often not available.
In this section, we evaluate our system for IS classification in the end-to-end setting. 
More specifically, we apply our IS assignment model described in Section \ref{sec:isAssignment} to four sets of system mentions separately.  The system mentions are predicted by four different approaches that we have compared in Section \ref{sec:mentionExtResult}.

Table \ref{tab:isnotes-extractedmention} reports the results of IS classification on ISNotes in the end-to-end setting. It is interesting to notice that although the first two baselines achieve similar results for mention extraction in terms of F-score (\emph{Syntactic NP} vs. \emph{Lee et al. (2011)} in table \ref{tab:mentionExt}), 
their results on IS classification are quite different on each IS category. More specifically, 
the mention detection component from \newcite{leeheeyoung11} performs much better on identifying \emph{old} and \emph{mediated/WorldKnowledge} mentions compared to the other simple baseline that predicts all NPs as mentions. This is in line with the fact that system mentions from \newcite{leeheeyoung11} are optimized to identify non-singletons for coreference resolution. In ISNotes, among all mentions that are the first mentions in  coreference chains, 
46.2\% are \emph{new} mentions, 23.2\% are \emph{m/worldKnowledge} mentions, and
18.0\% are \emph{m/syntactic} mentions.

In general, our proposed system outperforms the three baselines in all metrics for 
most
 IS categories by a large margin. 
We also notice that within our system, the IS classification results in the end-to-end setting degrade for all IS categories compared to the setting with gold mentions (Table \ref{tab:isnotes-extractedmention} vs. Table \ref{tab:isnotes-goldmention}). Particularly, 
the \emph{old} and \emph{mediated/bridging} categories have less performance degradation compared to other IS classes.

\begin{figure*}[t]
\begin{center}
\includegraphics[width=1.0\textwidth]{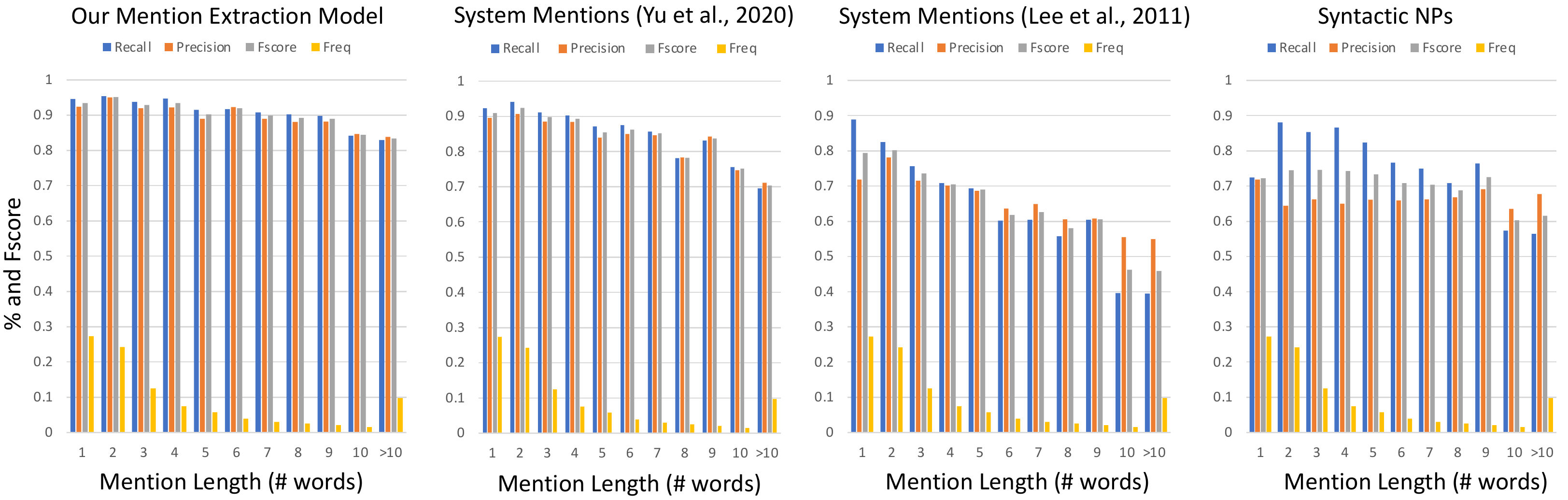}
\end{center}
\caption{Mention extraction performance for mentions of various lengths in different systems. ``Freq'' is the percentage of gold mentions with each length.}
\label{fig:mentionAnalysis}
\end{figure*}

\subsection{Results for Bridging Anaphora Recognition on BASHI and SciCorp}
To test the effectiveness of our system in a more realistic scenario, we apply our models trained on the whole ISNotes corpus to BASHI and SciCorp to recognize bridging anaphors. Note that BASHI also includes comparative anaphors as bridging anaphors. Therefore, we merge the predictions for \emph{mediated/bridging} and \emph{mediated/comparative} and treat them as bridging anaphors for BASHI. For SciCorp, we use a small list of determiners\footnote{The whole list of determiners we used to detect definite NPs is: \emph{\{the, this, that, these, those\}}.} to choose definite bridging anaphors from all predicted \emph{mediated/bridging} mentions since only definite noun phrases are annotated in SciCorp. 

Table \ref{tab:bridgingAna} shows the results of our system for bridging anaphora recognition on BASHI and SciCorp. We also list the results from a recent approach \cite{yu-poesio-2020-multitask} for these two corpora. \newcite{yu-poesio-2020-multitask}
propose a multi-task learning framework to identify bridging links and report results on 
BASHI and SciCorp using NPs as the predicted mentions. Note that the results from \newcite{yu-poesio-2020-multitask} are based on the models trained on the in-domain datasets using 10-fold cross-validation\footnote{The results on SciCorp from \newcite{yu-poesio-2020-multitask} may not be directly comparable to ours since it is unclear whether they filter out \emph{containing inferrable} mentions from bridging anaphors annotated in SciCorp.}. This is because SciCorp contains texts from a very different domain compared to ISNotes and BASHI.

In general, our system trained on ISNotes achieves competitive results to identify bridging anaphors on BASHI and SciCorp compared to \newcite{yu-poesio-2020-multitask}. We notice that our system achieves relatively high recall scores on both corpora (59.7\% on BASHI and 47.6\% on SciCorp). 

\begin{table}[t]
\begin{center}
\begin{tabular}{|l|c|c|c|}
\hline
 &R&P&F\\  \hline
\multicolumn{4}{|c|}{BASHI} \\ \hline
\emph{ Yu and Poesio (2020)}& 34.2&34.4&34.3\\ \hline
\emph{our system}&59.7&25.5&\textbf{35.8}\\ \hline
\multicolumn{4}{|c|}{SciCorp} \\ \hline
\emph{ Yu and Poesio (2020)}& 35.7&45.0&39.8\\ \hline
\emph{our system}&47.6&36.0&\textbf{41.0}\\ \hline
\end{tabular}
\end{center}
\caption{\label{tab:bridgingAna} Results of our system for bridging anaphora recognition on BASHI and SciCorp. ``\emph{Yu and Poesio (2020)}'' is the results from \newcite{yu-poesio-2020-multitask} which include gold coreferent anaphors for evaluation.
}
\end{table}

More interestingly, although our system is trained on a dataset consisting of news articles, it still can identify certain domain specific bridging anaphors in genetics and computational linguistic scientific papers from SciCorp, such as ``\emph{the underlying siRNA}'', ``\emph{the missing two Nucleotides}'', ``\emph{the target mRNA}", ``\emph{the previous optimization}",  ``\emph{the objective function}",  ``\emph{the next clause}'', and ``\emph{the most predictive features}''.
 \newcite{hou-2020-fine} states that some bridging can be indicated by referential patterns without world knowledge about the anaphor/antecedent NPs. 
 It seems that 
 our IS assignment model can capture some of these patterns and generalize them into 
 different domains.

\section{Analysis}
To better understand the strengths and weaknesses of our system, we provide both quantitative and qualitative analyses. In the following discussion, we use predictions from ISNotes.

\paragraph{Mention Extraction Performance.} The training dataset we generated offers strong signals for spans that correspond to entity mentions, including singletons and non-singletons.  In Figure \ref{fig:mentionAnalysis}, we show recall, precision, and F-score  for mentions of various lengths. In general, our mention extraction model (Figure \ref{fig:mentionAnalysis}: left) 
performs significantly better than the three baseline models for all length groups. 
Although there is some performance degradation for longer mentions containing more than ten words, the results of our model for this group are still quite strong with an F-score of 83.4. 

It is also worth noting that the results using the mention detection component from \newcite{leeheeyoung11} degrade gradually with the increase of the mention length (Figure \ref{fig:mentionAnalysis}: middle). 
We think this is because the system is optimized to identify non-singleton mentions for coreference resolution and most coreferent mentions are short. In ISNotes, 88.6\% of non-singleton mentions contain less than six words.  


\paragraph{Common Errors on Identifying Bridging.}
Although our system achieves substantial improvements for bridging anaphora recognition on ISNotes compared to previous work \cite{houyufang13b,yu-poesio-2020-multitask,hou-2020-fine}, detecting bridging anaphors still remains a challenging problem compared to other IS categories. We examine the confusion matrix of our IS assignment model based on the gold mentions. We find that the model mostly confuses bridging anaphors with \emph{new} mentions. 
The highest portion of recall errors is due to the fact that the model misclassifies 202 bridging anaphors as \emph{new} mentions, while most precision errors are 142 \emph{new} mentions being misclassified as \emph{mediated/bridging}. It seems that our model struggles to distinguish bridging anaphors from some generic \emph{new} mentions that  are relational nouns \cite{loebner85} (see Example \ref{ex:new1} and Example \ref{ex:new2}). 

\begin{examples}
\item \label{ex:new1}  \textbf{Rescue crews}, however, 
  gave up hope that others would be found alive under the collapsed roadway.
\item \label{ex:new2}  Lang is cutting \textbf{costs} and will attempt to operate the magazine with only subscription revenue.
\end{examples}

\section{Conclusions}
In this paper, we propose a system that addresses information status classification in the end-to-end setting for the first time. Our mention extraction model does not require any pruning process and can still achieve strong performance for longer spans. 
We show that our system outperforms other baselines for mention extraction and IS assignment in different settings. 
We further demonstrate that our system trained on ISNotes can be applied to identify 
bridging anaphors in different domains. 
In the future, we plan to further improve our algorithm 
for bridging anaphora recognition. 
Applying our system to detect mentions and anaphoricity for coreference resolution is another interesting avenue for future work.









\section*{Acknowledgments}
The author thanks the anonymous reviewers for their the valuable feedback.

\bibliographystyle{acl_natbib}
\bibliography{../bib/lit/lit,anthology,emnlp2021}

\newpage

\appendix
\section{Appendices}
\label{sec:appendix}
\subsection{Datasets}
The ISNotes corpus can be downloaded from \url{https://www.h-its.org/software/isnotes-corpus/}. The BASHI corpus can be downloaded from \url{https://www.ims.uni-stuttgart.de/en/research/resources/corpora/bashi/}. The SciCorp dataset can be downloaded from \url{https://www.ims.uni-stuttgart.de/en/research/resources/corpora/scicorp/}. Note that 
ISNotes needs to be integrated with the  original OntoNotes  ONF files, and BASHI needs to be merged with the OntoNotes CoNLL format annotations.

\subsection{Model Parameters}
Our system is developed based on Huggingface Transformers\footnote{https://github.com/huggingface/transformers}. 
The mention extraction model is trained for one epoch with a learning rate of $1e-5$ and a batch size of 32. The IS assignment model is trained for three epochs with a learning rate of $3e-5$ and a batch size of 32. Both models are initialized using pre-trained RoBERTa$_{LARGE}$ contextual embeddings. They have 24 transformer blocks, 1024 hidden units, 16 self-attention heads, and around 355M parameters.

During training and testing, the maximum text length is set to 128 tokens. The maximum text length is set to 256 tokens when we apply the system to SciCorp to recognize bridging anaphors. This is because SciCorp contains a few long sentences due to the noise from sentence splitting.

\begin{table*}[t]
\begin{footnotesize}
\begin{center}
\begin{tabular}{|p{1.8cm}|p{0.466cm}p{0.466cm}p{0.466cm}|p{0.466cm}p{0.466cm}p{0.466cm}|p{0.466cm}p{0.466cm}p{0.466cm}|p{0.466cm}p{0.466cm}p{0.466cm}|p{0.466cm}p{0.466cm}p{0.466cm}|}
\hline
 &  \multicolumn{12}{c|}{\emph{baselines}} &\multicolumn{3}{c|}{} \\ 
  & \multicolumn{3}{c|}{Hou et al.(2013)} &\multicolumn{3}{c|}{Hou et al.(2013)} & \multicolumn{3}{c|}{Hou (2016)} & \multicolumn{3}{c|}{Hou (2020)}& \multicolumn{3}{c|}{}\\ 
 &  \multicolumn{3}{c|}{\emph{collective}} &\multicolumn{3}{c|}{\emph{cascade collective}} & \multicolumn{3}{c|}{\emph{incremental LSTM }} & \multicolumn{3}{c|}{\emph{self-attention with}}& \multicolumn{3}{c|}{\emph{this work}}\\ 
&  \multicolumn{3}{c|}{} &\multicolumn{3}{c|}{} & \multicolumn{3}{c|}{} & \multicolumn{3}{c|}{\emph{BERT$_{LARGE}$}}& \multicolumn{3}{c|}{}\\ 
 & R & P & F & R & P & F& R & P & F& R & P & F& R & P & F\\ \hline
\hline
 &&&&&&&&&&&&&&& \\
 {old}&84.4&86.0&85.2&82.2&87.2&84.7&85.4&84.9&85.2&88.4&90.0&89.2&88.8&91.8&\textbf{90.3}\\
 {m/worldKnow.}&67.4&77.3&72.0&67.2&77.2&71.9&67.1&74.5&70.6&77.7&79.5&78.6&79.2&79.4&\textbf{79.3}\\
 {m/syntactic} & 82.2&81.9&82.0&81.6&82.5&82.0&80.8&81.9&81.4&83.7&81.1&82.4&84.7&83.5&\textbf{84.1}\\
 {m/aggregate}&64.5&79.5&71.2&63.5&77.9&70.0&67.8&84.6&75.3&80.1&79.3&\textbf{79.7}&76.8&77.5&77.1\\
 {m/function} & 67.7&72.1&69.8&67.7&72.1&69.8&64.6&76.4&70.0&73.4&85.5&79.0&90.6&81.7&\textbf{85.9}\\
 {m/comparative} & 81.8&82.1&82.0&86.6&78.2&82.2&77.9&83.1&80.4&90.5&86.7&\textbf{88.6}&87.7&88.1&87.9\\
 {m/bridging}& 19.3&39.0&25.8&44.9&39.8&42.2&15.7&32.3&21.1&51.0&54.5&52.7&54.1&59.9&\textbf{56.9}\\
 {new} & 86.5&76.1&81.0&83.0&78.1&80.5&87.2&74.8&80.5&86.6&85.2&85.9&88.8&85.8&\textbf{87.3}\\
 \hline
 acc &\multicolumn{3}{c|}{78.9} & \multicolumn{3}{c|}{78.6} & \multicolumn{3}{c|}{78.6} & \multicolumn{3}{c|}{83.7}&\multicolumn{3}{c|}{\textbf{85.1}}\\
\hline
\end{tabular}
\end{center}
\end{footnotesize}
\caption{Results of the IS assignment model compared to previous approaches on ISNotes based on gold mentions.}
\label{tab:results_all}
\end{table*}

\subsection{Computing Infrastructure and Running Time}
We carried out experiments in a computing cluster environment. We use a node with a v100 GPU and 32 GB RAM for all experiments. We perform 10-fold cross-validation on ISNotes. It takes around 120 minutes to train and test the mention extraction model for one fold. And it takes around 25 minutes to train and test the IS assignment model for one fold without any pruning.



\subsection{Results of previous studies }
\label{sec:moreisresults}
In this section we compare our IS assignment model to previous approaches which use  gold mentions as input. 
Table \ref{tab:results_all} shows the results of fine-grained IS classification of different models on ISNotes. Below is a short description of each model.

\paragraph{\emph{collective}.} \newcite{houyufang13b} applied collective classification to account for the linguistic relations among IS categories.
 They explored a wide range of features (34 in total), including a large number of lexico-semantic features (for recognizing bridging) as well as a couple of surface features and syntactic features.
 
\paragraph{\emph{cascaded collective}.} This is the cascading minority preference system for bridging anaphora recognition from \newcite{houyufang13b}. 

\paragraph{\emph{incremental LSTM}.} This is the attention-based LSTM model proposed by \newcite{houyufang16}. The model uses one-hot vectors to encode IS classes and predicts information status for all mentions of a document from left to right incrementally. 

\paragraph{\emph{self-attention with BERT$_{LARGE}$}.} \newcite{hou-2020-fine} treats IS classification as a sentence classification task based on carefully designed ``pseudo sentences''. It is fine-tuned on \emph{BERT$_{LARGE}$} and uses the hidden state of the ``[CLS]'' token for prediction.

\paragraph{\emph{this work.}} The model described in Section \ref{sec:isAssignment}.

\end{document}